\documentclass[a4paper,twoside]{article}

\usepackage{eucal}

\usepackage{xcolor}
\definecolor{mygray}{gray}{0.6}
\usepackage[flushleft]{threeparttable}
\newcommand*{\ditto}{---\texttt{"}---}
\usepackage{enumitem}
\usepackage[hyphens]{url}
\usepackage{hyperref}
\usepackage{graphicx}

\usepackage{epsfig}
\usepackage{subcaption}
\usepackage{calc}
\usepackage{amssymb}
\usepackage{amstext}
\usepackage{amsmath}
\usepackage{amsthm}
\usepackage{multicol}
\usepackage{pslatex}
\usepackage{apalike}
\usepackage{SCITEPRESS}     

\begin{document}

\title{On the Metrics and Adaptation Methods for Domain Divergences of sEMG-based Gesture Recognition}

\author{\authorname{Istv\'{a}n Ketyk\'{o}\orcidAuthor{0000-0003-4931-4580} and Ferenc Kov\'{a}cs\orcidAuthor{0000-0002-9777-0372}}
\affiliation{Applied ML Research Group, Nokia Bell Labs, Budapest, Hungary}
\email{\{istvan.ketyko, ferenc.2.kovacs\}@nokia-bell-labs.com}
}

\keywords{Machine Learning, Time-Series Modeling, sEMG/EMG, Divergence Metrics, Domain Adaptation.}

\abstract{We propose a new metric to measure domain divergence and a new domain adaptation method for time-series classification. The metric belongs to the class of probability distributions-based metrics, is transductive, and does not assume the presence of source data samples. The 2-stage method utilizes an improved autoregressive, RNN-based architecture with deep/non-linear transformation. We assess our metric and the performance of our model in the context of sEMG/EMG-based gesture recognition under inter-session and inter-subject domain shifts.}

\onecolumn \maketitle \normalsize \setcounter{footnote}{0} \vfill

\section{INTRODUCTION} \label{sec:introduction}
\noindent Machine Learning (ML) is widely used for several tasks with time-series and biosensor data such as for human activity recognition, electronic health records data-based predictions \cite{time-series_classification}, and real-time bionsensor-based decisions. Various classification goals are addressed related to electrocardiography (ECG) \cite{ECG-classification}, electroencephalography (EEG) \cite{EEG-classification,DOSE2018532}, and electromyograpy (EMG) \cite{2SRNN,10.1371/journal.pone.0206049,ICPR-2014-PatriciaTC,Du2017}.

Sensing hand gestures can be done by means of wearables or by means of image or video analysis of hand or finger motion. A wearable-based detection can physically rely on measuring the acceleration and rotations of our body parts (arms, hands or fingers) with Inertial Measurement Unit (IMU) sensors or by measuring the myo-electric signals generated by the various muscles of our arms or fingers with EMG sensors. Surface EMG (sEMG) records muscle activity from the surface of the skin which is above the muscle being evaluated. The signal is collected via surface electrodes.

\begin{figure}[t]
	\centerline{\includegraphics[width=\columnwidth,keepaspectratio]{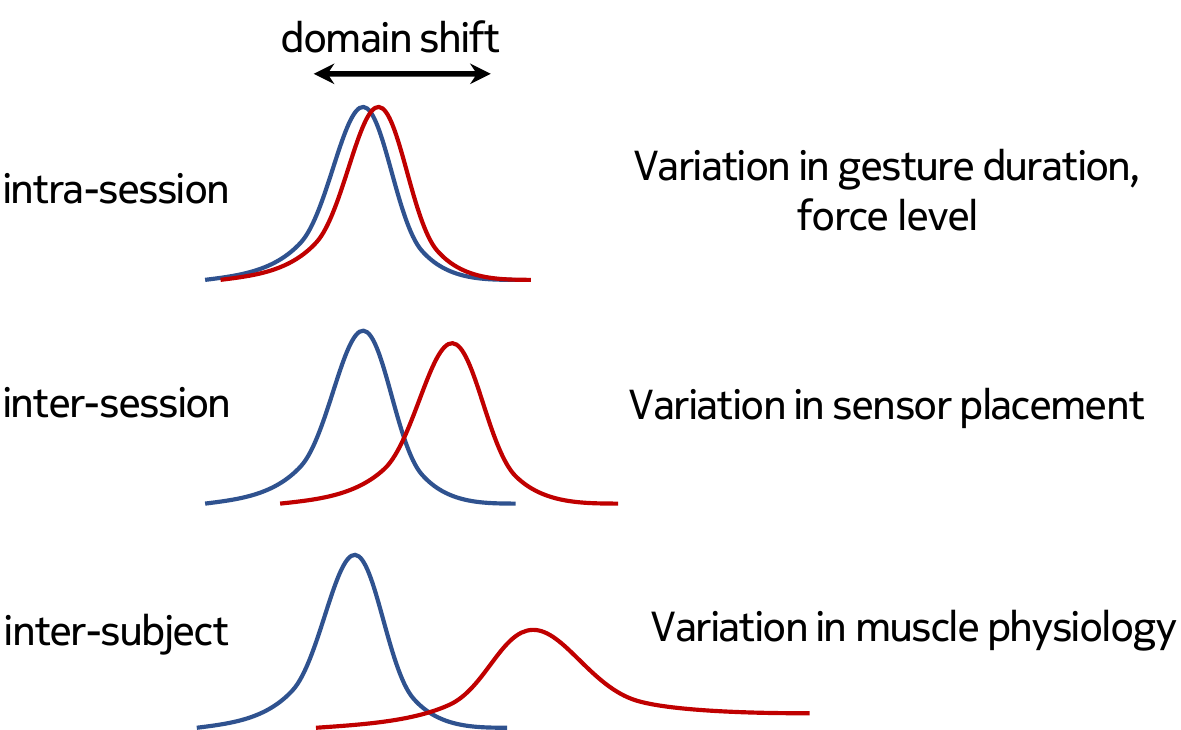}}
	\caption{Domain shift in case of different scenarios.}
	\label{figure:domainShift}
\end{figure}

We are interested in sEMG-sensor placement to the forearm and performing hand gesture recognition with ML. In this context, all ML prediction models suffer from inter-session and inter-subject domain shifts (see Figure~\ref{figure:domainShift}).

\begin{itemize}
	\item \emph{Intra session} scenario: the device is not removed, and the training and validation data are recorded together in the same session of the same subject. In this situation the gesture recognition accuracy is generally above $90\%$.
	\item \emph{Inter-session} scenario: the device is reattached, and the validation data is recorded separately in a new session of the same subject. Under this domain shift the validation accuracy degrades below $50\%$.
	\item \emph{Inter-subject} scenario: The validation data is on another subject. In this case, the validation accuracy degrades below $50\%$ as well.
\end{itemize}
Our focus is to investigate: 1) the metrics of these domain discrepancies, and 2) the adaptation solutions with special attention on those, which do not rely on source data samples.

This paper is organized as follows, Section~\ref{sec:related_work} provides a summary of ML, model risks, domains, domain divergences, and domain adaptation methods. Then our source data-absent metric and adaptation model is introduced in Section~\ref{sec:our_metric_and_method}. Next, we validate our approaches using publicly available sEMG datasets: the experimental setup and results are described in Section~\ref{sec:results}. Finally, we conclude and summarize our results.

\section{RELATED WORK} \label{sec:related_work}
\subsection{Machine Learning}

\noindent At the most basic level, ML seeks to develop methods for computers to improve their performance at certain tasks on the basis of observed data. Almost all ML tasks can be formulated as making inferences about missing or latent data from the observed data. To make inferences about unobserved data from the observed data, the learning
system needs to make some assumptions; taken together these assumptions constitute a model. The probabilistic approach to modeling uses probability theory to express all forms of uncertainty. Since any sensible model will be uncertain when predicting unobserved data, uncertainty plays a fundamental part in modeling. The probabilistic approach to modeling is conceptually very simple: probability distributions are used to represent all the uncertain unobserved quantities in a model (including structural, parametric and noise-related) and how they relate to the data. Then the basic rules of probability theory are used to infer the unobserved quantities given the observed data. Learning from data occurs through the transformation of the prior probability distributions (defined before observing the data), into posterior distributions (after observing data) \cite{Ghahramani2015}.

We define an \emph{input space} $\mathcal{X}$ which is a subset of $d$-dimensional real space $\mathbb{R}^d$. We define also a random variable $X$ with probability distribution $P(X)$ which takes values drawn from $\mathcal{X}$. We call the realisations of $X$ feature vectors and noted $x_i$.

A generative model describes the marginal distribution over $X$: $P_\Theta(X)$, where samples $x_i$ of $X$ are observed at learning time in a dataset $D$ and the probability distribution depends on some unknown parameter $\Theta$. A generative model family which is important for time-series analysis is the autoregressive one. Here, we fix an ordering of the variables $X_1,X_2, \ldots ,X_n$ and the distribution for the $i$-th random variable depends on the values of all the preceding random variables in the chosen ordering $X_1,X_2, \ldots ,X_{i-1}$ \cite{scheduled_sampling}. By the chain rule of probability, we can factorize the joint distribution over the $n$-dimensions as:
\begin{equation}
  \label{eq:autoregression}
  P_\Theta(X) = \prod_{i=1}^{n}{P_\Theta(X_i \mid X_1,X_2, \ldots, X_{i-1})}
\end{equation}
We define also an \emph{output space} $\mathcal{Y}$ and a random variable $Y$ taking values $y_i$ drawn from $\mathcal{Y}$ with distribution $P(Y)$. In the supervised learning setting $Y$ is conditioned on $X$ (i.e., $Y \sim P(Y \mid X)$), so the joint distribution $P(X,Y)$ is actually $P(X)P(Y \mid X)$.

A discriminative model is relaxed to the posterior conditional probability distribution: $P_\Theta(Y \mid X)$ and it reflects straight the discrimination/classification task with lower asymptotic errors than the generative models. This transductive learning setting has been introduced by Vapnik \cite{Ng:2001:DVG:2980539.2980648}.

For all modeling approaches, the learning is to fit their distributions over the observed variables $x_i$ in our dataset $D$. With other words (of a statistician), a good estimate of the unknown parameter $\Theta$ would be the value of $\Theta$ that maximizes the likelihood of getting the data we observed in our dataset $D$. Formally, the goal of the Maximum Likelihood Estimation (MLE) is to find the $\Theta$:
\begin{equation}
  \label{eq:MLE}
  \Theta = \operatorname*{argmax}_\Theta P_\Theta(D) = \operatorname*{argmax}_\Theta \sum_{x_i \in D} \log P_\Theta(x_i)
\end{equation}
\noindent Learning in neural networks is about solving optimisation problems. In case of probabilistic (and differentiable) cost functions, the backpropagation method \cite{backprop} is an estimator for MLE.

\subsection{Risk of Machine Learning Models}

We start with the general and probabilistic description then we move on to the discriminative models.

Let us introduce the concept of \emph{loss function}. A loss function $L(x_i, P_\Theta)$ measure the loss that a model distribution $P_\Theta$ makes on a particular instance $x_i$.
Our goal is to find the model $P_\Theta$ that minimizes the expected loss or \emph{risk}:
\begin{equation}
  \label{eq:risk}
  R(P_\Theta) = \mathbb{E}_{x_i \sim P(X)}[L(x_i, P_\Theta)]
\end{equation}
Note that the loss function which corresponds to MLE is the log loss $-\log{P_\Theta(x_i)}$. The risk evaluated on $D$ is the in-sample error or \emph{empirical risk}:
\begin{equation}
  \label{eq:empirical_risk}
  \hat{R}(P_\Theta) = \frac{1}{|D|}\sum_{x_i \in D}L(x_i, P_\Theta)
\end{equation}

The generalization gap is the $R(P_\Theta) - \hat{R}(P_\Theta)$ and our model intends to have small probability that this difference is larger than a very small value $\epsilon$.

In case of supervised learning setting, we have a paired dataset of observations $D=\{(x_1,y_1),\ldots,(x_m,y_m)\}$. The loss function becomes the conditional log-likelihood $L(x_i, y_i, P_\Theta)=-\log{P_\Theta(y_i \mid x_i)}$ and the risk is described as:
\begin{equation}
  \label{eq:risk_of_joint_variables}
  R(P_\Theta) = \mathbb{E}_{(x_i,y_i) \sim P(X,Y)}[L(x_i, y_i, P_\Theta)]
\end{equation}

Let us introduce the \emph{hypothesis space} $\mathcal{H}$ which is the set/class of predictors $\{h: \mathcal{X} \rightarrow \mathcal{Y}\}$. A hypothesis $h$ estimates the target function $f:\mathcal{X} \rightarrow \mathcal{Y}$ from $D$. The target function $f$ is a proxy for the conditional distribution $P(Y \mid X)$, such as: $y_i = f(x_i) + \zeta$, where $\zeta$ is the noise term. Substituting $h$ into Equation~\eqref{eq:risk_of_joint_variables} we get the transductive version of the risk:
\begin{equation}
  \label{eq:risk_of_joint_variables_with_h}
  R^{td}(h) = \mathbb{E}_{(x_i,y_i) \sim P(X,Y)}[L(y_i, h(x_i))]
\end{equation}
Substituting $h$ into Equation~\eqref{eq:empirical_risk}:
\begin{equation}
  \label{eq:empirical_risk_with_h}
  \hat{R}^{td}(h) = \frac{1}{|D|}\sum_{(x_i,y_i) \in D}L(y_i, h(x_i))
\end{equation}
In the transductive setting the generalisation gap can be quantified:
\begin{equation}
  \Pr[\underset{h \in \mathcal{H}}{\mathrm{sup}} \lvert R^{td}(h) - \hat{R}^{td}(h) \rvert > \epsilon]
  \label{eq:generalisation_gap}
\end{equation}

\subsection{Domain and Domain-Shift Concepts}
First, it is necessary to clarify what a domain is and what kind of domain discrepancies there can be. There are several good survey papers that describe this field deeply e.g., \cite{DBLP:journals/corr/abs-1812-11806}, \cite{DBLP:journals/corr/abs-1901-05335}, and \cite{DBLP:journals/corr/Csurka17}. In this paper, the domain adaptation-related problem statement and notations follow \cite{DBLP:journals/corr/abs-1812-11806}.

The problem statement is introduced from a classification point of view to simplify the definitions, but it can be generalized to other supervised machine learning task. A domain contains three elements: \emph{Input space} $\mathcal{X}$, \emph{Output space} $\mathcal{Y}$ and \emph{$P(X,Y) $} joint distribution over $\mathcal{X}$ and $\mathcal{Y}$.

Two domains are different if at least one of their above mentioned components are not equal. In case of domain adaptation the input spaces and output spaces of the domains are the same but the distributions are different. More general cases belong to different fields of transfer learning, a detailed taxonomy of transfer learning tasks can be found in \cite{TransferLaerningTaxonomy}. During domain adaptation there is a trained machine learning model on a so-called source domain (S) and there is an intent to apply it on a target domain (T). From this point onwards S and T in subscript refer to source and target domains.

Let us analyze the risk of a source classifier (Equation~\eqref{eq:risk_of_joint_variables_with_h}) on a target domain $T$ in the cross-domain setting:
\begin{align}
	\nonumber
	R_T(h) &= \sum_{y \in \mathcal{Y}} \int_{\mathcal{X}}L(h(x)\mid y)P_{T} (x,y)\,dx \\
		&= \sum_{y \in \mathcal{Y}} \int_{\mathcal{X}}L(h(x)\mid y)P_{S}(x,y) \frac{P_{T} (x,y)}{P_{S} (x,y)} \,dx
\end{align}
It can be seen that the ratio of the source and target joint distributions ($P_{T}(X,Y)/P_{S}(X,Y)$) defines the risk $R_T(h)$. The investigation of this ratio allows us to define domain shift cases \cite{Moreno-Torres:2012:UVD:2030819.2031236}: prior shift, covariate shift and concept shift.

In case of prior shift, the marginal distribution of the labels are different between the source domain and the target domain $P_{S}(Y)\neq P_{T}(Y)$, but the conditional distributions are equal $P_{S}(X \mid Y) = P_{T}(X \mid Y)$. Typical example for prior shift: the symptoms of a disease are usually population independent but the distribution of the diseases is population dependent. These conditions allow us to simplify the risk:
\begin{align}
	\nonumber
	R_T(h) &= \sum_{y \in \mathcal{Y}} \int_{\mathcal{X}}L(h(x)\mid y)P_{S}(x,y) \frac{P_{T} (x,y)}{P_{S} (x,y)} \,dx \\
		\nonumber
		&= \sum_{y \in \mathcal{Y}} \int_{\mathcal{X}}L(h(x)\mid y)P_{S}(x,y) \frac{P_{T} (x \mid y)P_{T}(y)}{P_{S} (x \mid y) P_{S}(y)} \,dx \\
		&= \sum_{y \in \mathcal{Y}} \int_{\mathcal{X}}L(h(x)\mid y)P_{S}(x,y) \frac{P_{T} (y)}{P_{S} (y)} \,dx 
\end{align}
This means, that the complete labeled dataset from the target domain is not needed but the estimation of the marginal distribution of the labels $P_T(Y)$ is needed on the target domain.

Covariate shift is a well-studied domain shift, for further reference see \cite{DBLP:journals/corr/abs-1812-11806}. It is defined as follows: the posterior distributions are equivalent, this means $P_{T}(Y \mid X) = P_{S}(Y \mid X)$, but the marginal distributions of the samples are different $P_{S}(X) \neq P_{S}(X)$. The typical cause of covariate shift is the sample selection bias. Only the sample distributions determine the risk:
\begin{align}
 	\nonumber
	R_T(h) &= \sum_{y \in \mathcal{Y}} \int_{\mathcal{X}}L(h(x)\mid y)P_{S}(x,y) \frac{P_{T} (x,y)}{P_{S} (x,y)} \,dx \\
		\nonumber 
		&= \sum_{y \in \mathcal{Y}} \int_{\mathcal{X}}L(h(x)\mid y)P_{S}(x,y) \frac{P_{T} (y \mid x)P_{T}(x)}{P_{S} (y \mid x) P_{S}(x)} \,dx \\ 
		&= \sum_{y \in \mathcal{Y}} \int_{\mathcal{X}}L(h(x)\mid y)P_{S}(x,y) \frac{P_{T} (x)}{P_{S} (x)} \,dx 
\end{align}

In case of concept shift, the marginal distributions of input vectors are similar on both source and target domains $P_{S}(X) = P_{T}(X)$, on the other hand, the posterior distributions differ from each other $P_{T}(Y \mid X) \neq P_{S}(Y \mid X)$. Usually, non-stationary environment causes this data drift \cite{Widmer:1996:LPC:226791.226798}. It is not possible to simplify significantly the cross-domain risk and the domain adaptation cannot be done without labeled target data:
\begin{align}
	\nonumber
	R_T(h) &= \sum_{y \in \mathcal{Y}} \int_{\mathcal{X}}L(h(x)\mid y)P_{S}(x,y) \frac{P_{T} (x,y)}{P_{S} (x,y)} \,dx \\
		\nonumber
		&= \sum_{y \in \mathcal{Y}} \int_{\mathcal{X}}L(h(x)\mid y)P_{S}(x,y) \frac{P_{T} (y \mid x)P_{T}(x)}{P_{S} (y \mid x) P_{S}(x)} \,dx \\
		&= \sum_{y \in \mathcal{Y}} \int_{\mathcal{X}}L(h(x)\mid y)P_{S}(x,y) \frac{P_{T} (y \mid x)}{P_{S} (y \mid x)} \,dx
\end{align}

In general, none of the above mentioned assumption is valid, thus it is not possible to simplify the risk on target domain. The differing posterior distributions cause the major domain shift related issues. The optimal transport approach assumes that there is transport $t()$ that satisfies $P_{T}(Y \mid t(X)) = P_{S} (Y \mid X)$ \cite{7586038}. Finding this transportation map is intractable but it is possible to relax it to a simpler optimization problem, where $t()$ is estimated via a Wasserstein distance minimization between the two domains \cite{7586038,DBLP:journals/corr/abs-1901-05335}. 

\subsection{Divergence Metrics and Theoretical Bounds}
As the input space and output space are common in case of domain adaptation, the distance and divergence metrics of the distributions can measure and quantify the domain discrepancies. We elaborate the most common metrics in the field of domain adaptation. 

The Kullback-Leibler divergence \cite{Cover:1991:EIT:129837} is a well-known information theory-based metrics between two distributions. It measures the relative entropy between two distributions. One of its main disadvantage is, that it is difficult to calculate it from samples in some cases \cite{Ben-David2010}.
\begin{align}
	\label{eq:KL}
	\nonumber
	D_{KL}(P_{S} \| P_{T}) &=\int_{\mathcal{X} \times \mathcal{Y}}P_{S}(x,y)\log\left(\frac{P_{S}(x,y)}{P_{T}(x,y)}\right)dxdy \\
	&= \sum_{y \in \mathcal{Y}} \int_{\mathcal{X}}P_{S}(x,y)\log\left(\frac{P_{S} (x,y)}{P_{T} (x,y)}\right)dx
\end{align}
In general, the KL divergence is an asymmetric metric as $D_{KL}(P_{S} \| P_{T}) \neq D_{KL}(P_{T} \| P_{S})$. A commonly used symmetric version is the Jensen-Shannon divergence \cite{lin1991divergence}. It measures the total divergence from the average divergence.
\begin{align}
	\nonumber
	M &= \frac{P_{S} + P_{T}}{2} \\	
	D_{JS}(P_{S} \| P_{T}) &= \frac{1}{2} \left(D_{KL}(P_{S} \|M) + D_{KL}(P_{T} \|M)\right)
\end{align}
 
The origin of the Wasserstein distance is the optimal transport problem: a distribution of mass should be transported to another distribution of mass with minimal cost. Usually Wassersten-1 distance is used with the Euclidean distance \cite{WGAN}.
\begin{equation}
	\label{eq:W}
	D_W(P_{S}, P_{T}) = \inf_{\gamma \in \Pi(P_S,P_T)}  \mathbb{E}_{(s,t) \sim \gamma}\left[\|s-t\|\right]
\end{equation}
Where $\Pi(P_S,P_T)$ is the set of all joint distributions with marginals $P_S$ and $P_T$. This distance metric allows us to construct a continous and differentiable loss function \cite{WGAN}. In case of domain adaptation, this distance is calculated between the marginal distributions $P_S(X)$ and $P_T(X)$ to get a tractable problem \cite{DBLP:journals/corr/abs-1812-11806}.

The H divergence allows to find upper bound to cross-domain risk \cite{Kifer:2004:DCD:1316689.1316707,Ben-David:2006:ARD:2976456.2976474,Ben-David2010}. The definitions and formulas are provided for binary classification because of simplification, but they can be generalized to multi-class problems, as well:
\begin{align}
	\nonumber	
	D_{H}(P_S,P_T) =& \\
	& \hspace{-15mm}
	2 \underset{h \in \mathcal{H}}{\mathrm{sup}} \lvert \Pr_{x \sim P_S}(h(x)=1) - \Pr_{x \sim P_T}(h(x)=1) \rvert
\end{align}
\cite{Ben-David:2006:ARD:2976456.2976474} provide two different techniques to estimate H divergence: from finite sample and from empirical risk of domain classifier. If the hypothesis space is symmetrical, the empirical H divergence can be calculated form finite samples of the source and target domains:
\begin{align}
	\label{eq:D_hat_H}
	\nonumber 
	 & \hat{D}_H(P_S,P_T) \ = \\
	 & 2\Bigg(\! 1\! -\! \min_{h \in \mathcal{H}} \bigg[\frac{1}{m} \sum_{i=1}^m I[h(x_{S,i})\!=\!1]  + \frac{1}{m'} \sum_{i=1}^{m'}I[h(x_{T,i})\!=\!0]\bigg]\! \Bigg)   
\end{align}
where $I$ is an indicator function which gives $1$ if predicate is correct, otherwise $0$. For the computation of $\hat{D}_H(P_S,P_T)$ during the minimization, the whole hypothesis space $\mathcal{H}$ must be tackled. \cite{Ben-David:2006:ARD:2976456.2976474} introduced an approximation to empirical H divergence, which is called Proxy-A Distance:
\begin{equation}
	\hat{D}_A(P_S, P_T) = 2(1-2\hat{R}_{S/T}),
\end{equation}
where $\hat{R}_{S/T}$ is the empirical risk of a linear domain classifier, which is trained (in a supervised fashion) to distinguish the source and target domains. 

The cross-domain risk can be estimated by the empirical H divergence:
\begin{equation}
	\label{eq:R_T_upper_bound}
	R_T(h)  \leq R_S(h) + R_{S+T}(h^*) + \hat{D}_H(P_S,P_T) + C(\mathcal{H}),
\end{equation}
where $C(\mathcal{H})$ is a complexity measure of hypothesis space, $R_{S+T}(h^*)$ is the risk of the so-called \emph{single good hypothesis}. The $h^*$ is the best classifier that can generalize on both domains:
\begin{align}
	\nonumber
	R_{S+T} = R_S(h) + R_T(h)\\
	h^* = \underset{h \in \mathcal{H}}{\mathrm{argmin}}(R_{S,T}(h))
\end{align}
The minimization of H divergence gives better result, however the the risk of single good hypothesis can ruin the performance of the domain adaptation.  In other words, if there is no single good hypothesis, the domains are too far from each other to build an efficient domain adaptation. 

\subsection{Domain Adaptation Techniques} \label{ssec:domain_adaptation_techniques}

All the adaptive learning strategies focus on identifying how to leverage the information coming from both the source and target domains. Incorporating exclusively the target domain
information is disadvantegous because sometimes there is no labelled targed data at all, or typically the amount of the labelled target data is small. Building on information present in the source domain and adapting that to the target is generally expected to be the superior \cite{ICPR-2014-PatriciaTC} solution.

We make a split in the viewpoint of source-sample availability at DA time. We discuss separately methods that assume source sample availability and methods that do not, first generally, and later in the context of sEMG-based gesture recognition.

\subsubsection{Source Data-based} \label{sssec:source_data_based}

The majority of the approaches incorporate the unlabeled source data samples at DA time. Cycle Generative Adversarial Network (Cycle-GAN) \cite{CycleGAN} is a state-of-the-art deep generative model which is able to learn and implicitly represent the source and target distributions to pull them close together  in an adversarial fashion. It is composed of two Generative Adversarial Networks (GANs) \cite{GAN} and learns two mappings ($A:T \rightarrow S$ and $B:S \rightarrow T$) to achieve the cycle-consistency between the source and target distributions ($A(B(S)) \approx S$ and $B(A(T)) \approx T$) via the minimax game.

Besides the GANs, the autoassociative Auto Encoder (AE) models are capable of building domain-invariant representations in their latent space. The non-linear Denoising AE (DAE) \cite{denoising_AE_based_DA} builds strong representation of the input distribution with the help of mastering to denoise the input (augmented with noise or corruptions). As a side effect, the multi-domain input ends up with a domain-invariant latent representation in the model. Inspired by the DAE, a linear counterpart: the Marginalized Denoising AE (mDA) \cite{Chen:2012:MDA:3042573.3042781} has been proposed to keep the optimization convex with closed-form solution and achieve orders-of-magnitude faster computation (at the expense of the representation power is limited be to linear).

Data augmentation with marginalized corruptions has been studied for the transductive learning setting \cite{Ng:2001:DVG:2980539.2980648} also: the Marginalized Corrupted Features (MFC) classifier \cite{DBLP:journals/corr/MaatenCTW14} has strong performance in case of validation under domain shift. In particular, as the corrupting distribution may be used to shift the data distribution in the source domain towards the data distribution in the target domain - potentially, by learning the parameters of the corrupting distribution using maximum likelihood.

In the transductive learning setting a classifier can be explicilty guided to learn a domain-invariant representation of the conditional distribution of class labels $P(Y \mid X)$ among two domains. Domain-Adversarial Neural Network DANN) \cite{Ganin:2016:DTN:2946645.2946704} adversarially connects a binary domain classifier into the neural network directly exploiting the idea exhibited by Equation~\eqref{eq:R_T_upper_bound}.

The binary domain classifier of the DANN \cite{Ganin:2016:DTN:2946645.2946704} and the mDA \cite{Chen:2012:MDA:3042573.3042781} have been paired in \cite{clinchant-etal-2016-domain} to get domain-adaptation regularization for the linear mDA model. Hence, the mDA has been explicitly guided to develop a latent representation space which is domain-invariant. Linear classifiers built in that latent space have have had comparable performance results in several image classification tasks.

The 2-Stage Weighting framework for Multi-Source Domain Adaptation (2SW-MDA) and the Geodesic Flow Kernel (GFK) methods in \cite{ICPR-2014-PatriciaTC} tackle inter-subject DA for sEMG-based gesture recognizers. In 2SW-MDA all the data of each source subject are weighted and combined with the target subject samples with a linear supervised method; for GFK the source and target data are embedded in a low-dimensional manifold (with PCA) and the geodesic flow is used to reduce the domain shift when evaluating the cross domain sample similarity.

\subsubsection{Source Data-absent} \label{sssec:source_data_absent}

The overwhelming majority of existing domain adaptation methods makes an assumption of freely available source domain data. An equal access to both source and target data makes it possible to measure the discrepancy between their distributions and to build representations common to both target and source domains. In reality, such a simplifying assumption rarely holds, since source data are routinely a subject of legal and contractual constraints between data owners and data customers \cite{Chidlovskii:2016:DAA:2939672.2939716}.

Despite the absence of available source samples it is still possible to rely on: 1) statistical information of the source retrieved in advance, 2) model(s) trained on the source data.

CORrelation ALignment (CORAL) \cite{coral} minimizes domain shift by aligning the second-order statistics of source and target distributions, without requiring any target labels. In contrast to subspace manifold methods (e.g., \cite{fernando2013SA}), it aligns the original feature distributions of the source and target domains, rather than the bases of lower-dimensional subspaces. CORAL performs a linear whitening transformation on the source data then a linear coloring transformation (based on the second-order statistics of the target data). If the statistical parameters of the source data are retrieved in advance of the DA then it can be considered as a source data-absent method.

Adaptive Batch Normalization (AdaBN) (which is an approximation of the whitening transformation, usually applied in deep neural networks) is utilised for DA in \cite{Du2017} for sEMG-based gesture classification. Furthermore, it builds upon the deep Convolutional Neural Network (CNN) architecture to extract spatial information from the high-density sEMG sensor input. However, it is not modeling the possible temporal information in the time-series data. Apart from that, it has state-of-the unsupervised DA performance which has been validated under inter-session and inter-subject domain shifts on several datasets.

\cite{fernando2013SA} introduces (linear) subspace alignment between the source and target domains with PCA. In the common linear subspace classifiers can be trained with comparable performance. The alignments (i.e., the PCA transformations) are learned on the source and target data, respectively. If the source alignment is learned (as a model of the source) in advance of the DA then it can be considered as a source data-absent method.

\cite{farshchian2018adversarial} introduces the Adversarial Domain Adaptation Network (ADAN) with an AE (trained on the source data) for Brain-Machine Interfaces (BMIs). With the representation power of the AE it is possible to capture the source distribution then continously align the shifting target distributions back to it. ADAN is trained in an adversarial fashion with an Energy-based GAN architecture \cite{EBGAN} where the "energy" is the reconstruction loss of the AE, and the domain shifts are represented as the residual-loss distributions of the AE. ADAN learns via the minimax game to pull the target residual distributions to those of the source.
 
In the transductive learning setting \cite{Ng:2001:DVG:2980539.2980648} there are several source data-absent DA approaches building on the pre-trained source classifier(s).

The Transductive Doman Adaptation (TDA) in \cite{Chidlovskii:2016:DAA:2939672.2939716} utilizes the representation capabilities of the mDA \cite{Chen:2012:MDA:3042573.3042781} to (linearly) adapt the output of a trained source classifier to the target domain. TDA performs unsupervised DA in closed form without the presence of any extra source information.

The transductive Multi-Adapt and the Multi-Kernel Adaptive Learning (MKAL) in \cite{ICPR-2014-PatriciaTC} both tackle the inter-subject DA for sEMG-based gesture recognizers by the adaptation of trained source classifiers. In Multi-Adapt, an SVM is learned from each source and used as reference (resulted by a convex optimization) when performing supervised learning on the target dataset. In MKAL each SVM source classifier predicts on the target samples and the scores are used as extra input features for the learning of the gesture classifier on the target dataset. Multi-Adapt and MKAL have had comparable performance at that time even though these models do not capture the available temporal information in the time-series data.

\cite{DOSE2018532} builds a BMI and investigates DA for multi-variate EEG time-series data classification. The time-series classification of the multi-variate EEG signals is a very similar challenge to the multi-variate sEMG signals. \cite{DOSE2018532} captures both the spatial and temporal correlations in the data with a CNN architecture. However, the DA is about supervised fine-tuning of all the model parameters on the target subject (such as \cite{pmlr-v32-donahue14}) which is suboptimal as highlighted by \cite{Du2017,2SRNN}.

The 2-Stage Recurrent Neural Network (2SRNN) model for sEMG gesture recognition and DA in \cite{2SRNN} can be viewed as the deep neural, autoregressive modeling analogy of the MKAL \cite{ICPR-2014-PatriciaTC}. It utilizes a trained source classifier and performs supervised DA to the target (session or subject) via learning a linear transformation between the domains. The transformation is then applied to the input (samples coming from the target). Learning is on the divergent (inter-session or inter-subject) domains via the backpropagation \cite{backprop} of the classifier's cross-entropy loss to its DA layer (which is a linear readout layer of the input). The size of its DA layer is less than $1\%$ of the overall 2SRNN (in terms of the trainable parameters) so it achieves fast computation of the DA, and the 2SRNN has the state-of-the-art performance in inter-session and inter-subject domain shift validations.

\section{OUR DIVERGENCE METRIC AND ADAPTATION METHOD} \label{sec:our_metric_and_method}

\noindent We provide a sequential, source data-absent, transductive, probability-based divergence metric and DA method as well. First, we introduce the RNN architecture for temporal modeling, then the source data-absent and transductive 2SRNN model in details.

\subsection{Recurrent Neural Network} \label{ssec:rnn}

\begin{figure}[tbp]
	\centerline{\includegraphics[width=0.8\columnwidth,keepaspectratio]{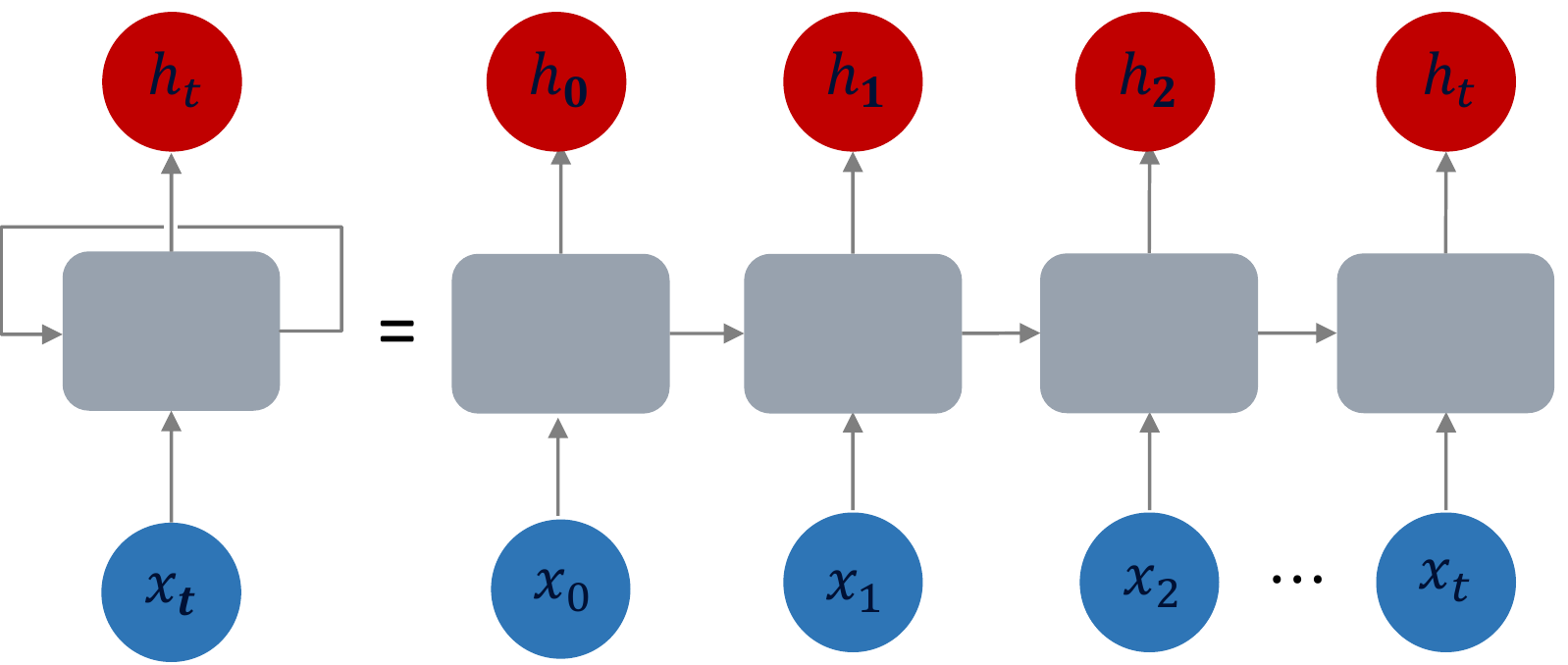}}
	\caption{RNN in compact and unrolled representations.}
	\label{figure:RNN}
\end{figure}

Recurrent Neural Network (RNN) \cite{RNN} is an autoregressive neural network architecture in which there are feedback loops in the system. Feedback loops allow processing the previous output with the current input, thus making the network stateful, being influenced the earlier inputs in each step (see Figure~\ref{figure:RNN}). A hidden layer that has feedback loops is also called a recurrent layer. The mathematical representation of a simple recurrent layer can be seen in Equation~\eqref{eq:rnn}:

\begin{equation} \label{eq:rnn}
\begin{aligned}
& \mathbf{h}_{t} = \sigma_{h} (\mathbf{W}_{x}\mathbf{x}_{t}+\mathbf{W}_{h}\mathbf{h}_{t-1}+\mathbf{b}_{t}) \\
& \mathbf{y}_{t} = \sigma_{y} (\mathbf{W}_{y}\mathbf{h}_{t}+\mathbf{b}_{y})
\end{aligned}
\end{equation}

The hidden state $\mathbf{h}_{t}$ depends on the input $\mathbf{x}_{t}$ and the previous hidden state $\mathbf{h}_{t-1}$. There is a non-linear dependency (via the $\sigma()$ wrapper) between them.

\begin{figure}[tbp]
\centerline{\includegraphics[width=200px,keepaspectratio]{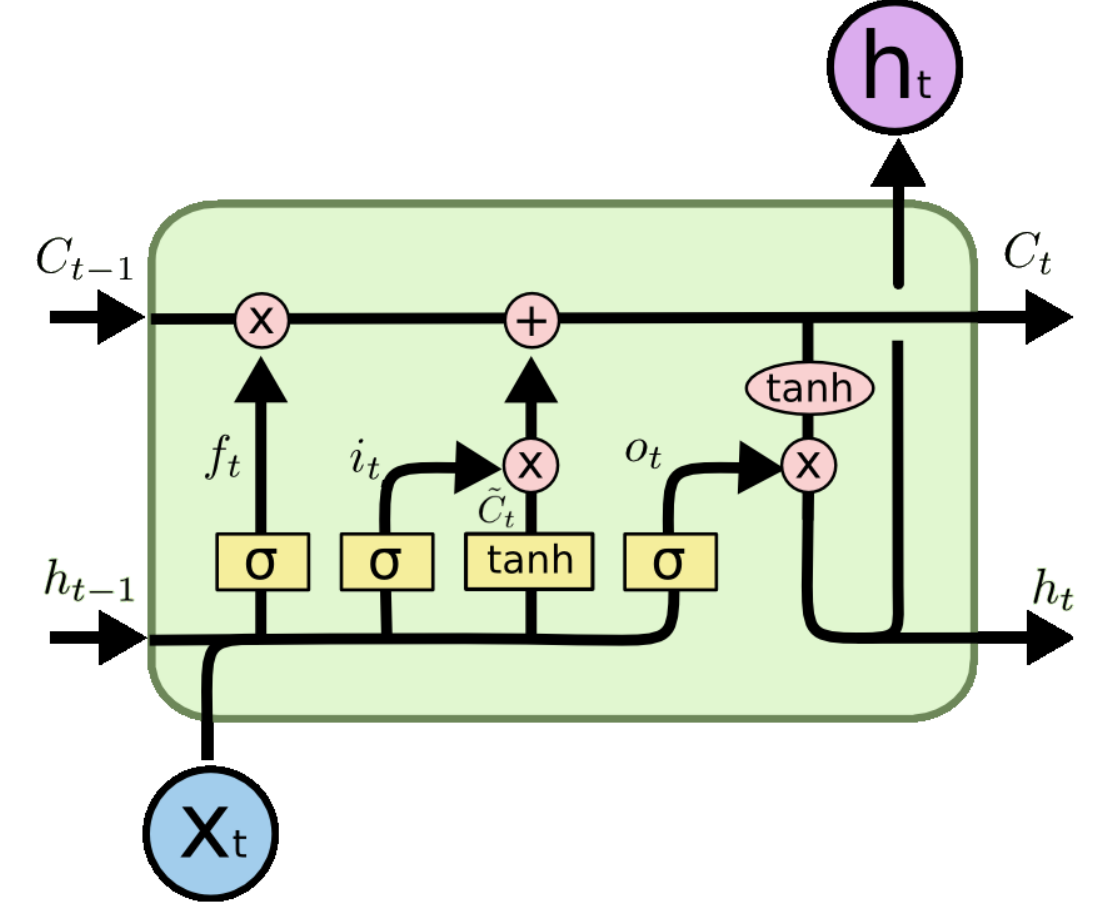}}
\caption{LSTM cell architecture \cite{colahweb}.}
\label{lstm_arch}
\end{figure}

However, regular RNNs suffer from the vanishing or exploding gradient problems which means that the gradient of the loss function decays/rises exponentially with time, making it difficult to learn long-term temporal dependencies in the input data \cite{razvan}. Long Short Term Memory (LSTM) recurrent cells have been proposed to solve these \cite{LSTM}.
\begin{equation} \label{eq:lstm}
\begin{aligned}
& \mathbf{f}_{t} = \sigma( \mathbf{W}_{f} \cdot [ \mathbf{h}_{t-1}, \mathbf{x}_{t} ] + \mathbf{b}_{f} )  \\
& \mathbf{i}_{t} = \sigma( \mathbf{W}_{i} \cdot [ \mathbf{h}_{t-1}, \mathbf{x}_{t} ] + \mathbf{b}_{i} ) \\
& \mathbf{\widetilde{C}}_{t} = \tanh( \mathbf{W}_{C} \cdot [ \mathbf{h}_{t-1}, \mathbf{x}_{t} ] + \mathbf{b}_{C} ) \\
& \mathbf{C}_{t} = \mathbf{f}_{t} \ast \mathbf{C}_{t-1} + \mathbf{i}_{t} \ast \mathbf{\widetilde{C}}_{t} \\
& \mathbf{o}_{t} = \sigma( \mathbf{W}_{o} \cdot [ \mathbf{h}_{t-1}, \mathbf{x}_{t} ] + \mathbf{b}_{o} ) \\
& \mathbf{h}_{t} = \mathbf{o}_{t} \ast \tanh(\mathbf{C}_{t})
\end{aligned}
\end{equation}
LSTM units contain a set of (learnable) gates that are used to control the stages when information enters the cell (input gate: $\mathbf{i}_{t}$), when it is output (output gate: $\mathbf{o}_{t}$) and when it is forgotten (forget gate: $\mathbf{f}_{t}$) as seen in Equation~\eqref{eq:lstm}. This architecture allows the neural network to learn longer-term dependencies because it learn also how to incorporate an additional information channel $\mathbf{C}_{t}$. In Figure~\ref{lstm_arch} yellow rectangles represent a neural network layer, circles are point-wise operations and arrows denote the flow of data. Lines merging denote concatenation (notation of $[]$ in Equation~\eqref{eq:lstm}), while a line forking denote its content being copied and the copies going to different locations.

For autoregressive modeling of time-series data, RNN with LSTM cells is widely adopted \cite{10.1371/journal.pone.0206049,2SRNN}.

\subsection{2-Stage Recurrent Neural Network-based Domain Divergence Metric} \label{ssec:our_divergence_metric}

Similarly to ADAN \cite{farshchian2018adversarial}, we build a source data-absent, probability-based divergence metric on the validation loss of the source model to measure domain shifts. In ADAN, the distribution of the residual loss (of the AE) is incorporated to express the divergence of target distributions from the one of source. However, we follow the transductive learning setting and directly take the (cross-entropy) loss of the source classifier (exhibiting Equation~\eqref{eq:risk_of_joint_variables}). Our source classifier is a sequential model (i.e., built on the autoregressive RNN architecture to have temporal modeling capabilites). For this task, we utilise the sequence classifier of the 2SRNN architecture \cite{2SRNN}.

\begin{figure}[tbp]
	\centerline{\includegraphics[width=\columnwidth,keepaspectratio]{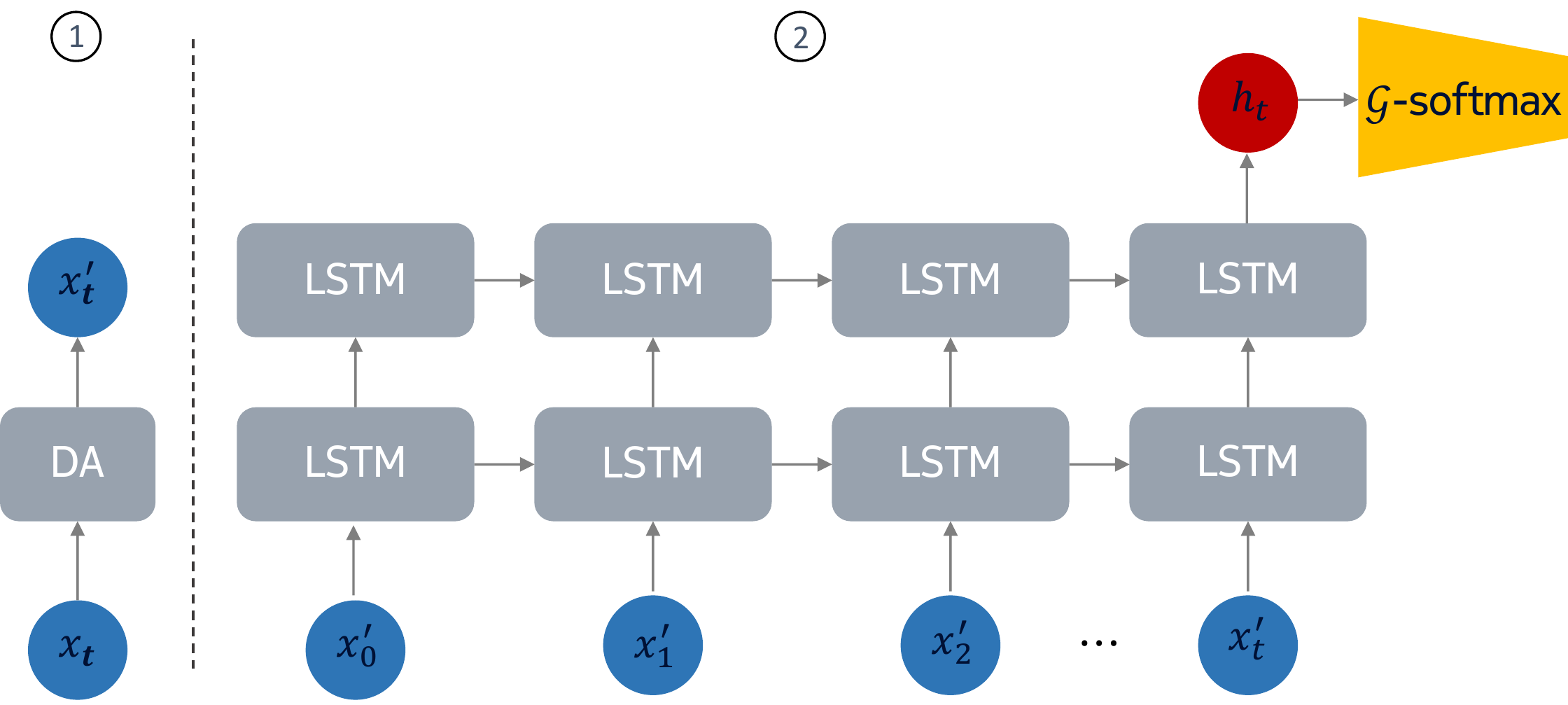}}
	\caption{The 2SRNN architecture: \newline 1) DA component, 2) Sequence classifier.}
	\label{figure:2SRNN}
\end{figure}

The sequence classifier of 2SRNN (visualised as block 2 in Figure~\ref{figure:2SRNN}) is a deep stacked RNN with the many-to-one setup followed by a $G$-way fully-connected layer ($G$ is the number of gestures to be recognized) and a softmax transformation at the output. The sequence classifier is directly modeling the conditional distribution of $P(Y \mid X)$, where $Y$ belongs to a categorical distribution with $G$ (gesture) classes. Learning is via the categorical cross-entropy loss (of the ground truth and the predicted $Y$):

\begin{equation}
  \label{eq:categorcial_cross_entropy}
  \mathcal{L}_{\mbox{\tiny{cross-entropy}}} = -\sum_{g \in G} {\textrm{I}_{g} \log P_{\Theta,g}},
\end{equation}

\noindent where $I_g$ is the indicator function whether class label $g$ is the correct classification for the given observation and $P_{\Theta,g}$ is the predicted probability that the observation is of class $g$.

For the divergence measure of distributions (between the source and target domains), we take the categorical cross-entropy losses of the sequence classifier in the following way: the classifier is trained on the source distribution then evaluated on a target one. Hence, the resulting $\mathcal{L}_{\mbox{\tiny{cross-entropy}}}$ expresses the domain shift in the loss space of the two domains. The cross-entropy between $P_S$ and $P_T$:
\begin{equation}
  \label{eq:cross_entropy}
  H(P_S, P_T) = H(P_S) + D_{KL}(S||T).
\end{equation}
\noindent The $\mathcal{L}_{\mbox{\tiny {cross-entropy}}}$ expresses the empirical $H(P_S, P_T)$ by the model. A valid source classifier is expected to model the source with the entropy of $H(P_S) \approx 1/G$, so in fact the cross-entropy $H(P_S, P_T)$ captures the actual Kullback-Leibler divergence among $P_S$ and $P_T$ (Equation~\eqref{eq:KL}).

Furthermore, let $\mu_S$ and $\mu_T$ be the corresponding means of $P_S$ and $P_T$. We measure the dissimilarity between these two distributions by a lower bound to the Wasserstein distance (Equation~\eqref{eq:W}), provided by the absolute value of the difference between the means \cite{BEGAN}:
\begin{equation}
  \label{eq:approx_wasserstein_distance}
  D_W(P_S, P_T) \geq |\mu_S - \mu_T|.
\end{equation}
The difference of the empirical means with the $\mathcal{L}_{\mbox{\tiny {cross-entropy}}}$ approximates Equation~\eqref{eq:approx_wasserstein_distance}.

\subsection{2-Stage Recurrent Neural Network-based Domain Adaptation} \label{ssec:our_DA}

\begin{figure}[tbp]
	\centerline{\includegraphics[width=\columnwidth,keepaspectratio]{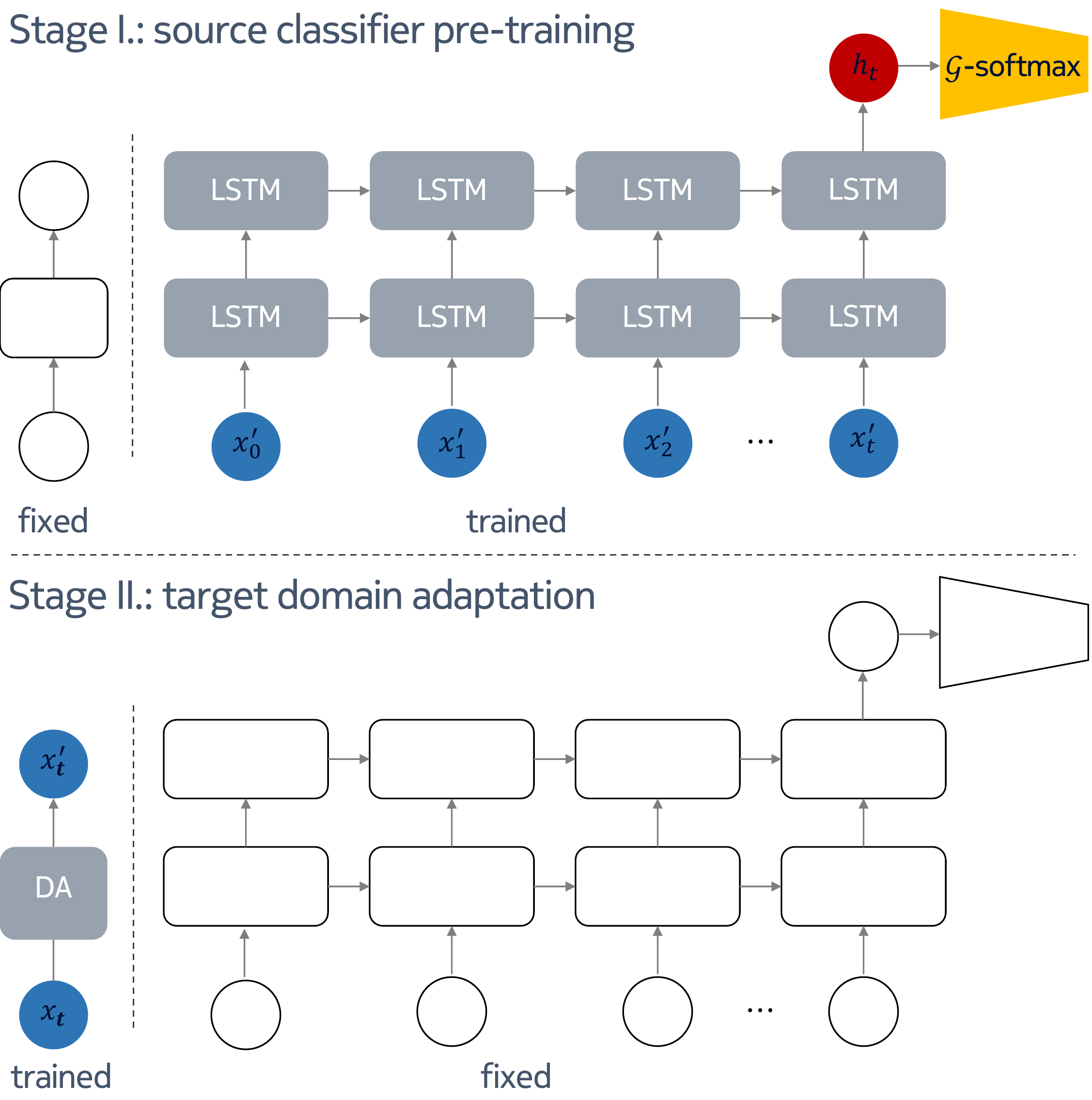}}
	\caption{The 2SRNN method: \newline I) Source classifier pre-training, II) Domain adaptation.}
	\label{figure:2SRNN-DA}
\end{figure}

We build a source data-absent, probability distribution of $\mathcal{L}_{\mbox{\tiny{cross-entropy}}}$-driven DA. \cite{2SRNN} implements a linear version (L-2SRNN), we extend it to a deep, non-linear one, and name it the Deep 2SRNN (D-2SRNN). Generally, the DA is applied to the input of the sequence classifier at each timestamp $t$ (visualised as block 1 in Figure~\ref{figure:2SRNN}). L-2SRNN learns the weights of a linear transformation:
\begin{equation}
  \label{eq:L-2SRNN}
  \mathbf{x'_t}=\mathbf{M}\mathbf{x_t} + \mathbf{b}.
\end{equation}
\noindent D-2SRNN learns the weights of chained non-linear transformations:
\begin{equation}
  \label{eq:D-2SRNN}
  \mathbf{x'_t} = \sigma(\mathbf{M_2} \cdot \sigma(\mathbf{M_1}\mathbf{x_t} + \mathbf{b_1}) + \mathbf{b_2}).
\end{equation}
\noindent Figure~\ref{figure:2SRNN-DA} presents the two consecutive stages of the DA process:

\begin{enumerate}[label=\Roman*)]
	\item The DA component initially is the identity transformation, and the weights of it are frozen. The sequence classifier is trained from scratch on the labelled source dataset.
	\item The weights of the sequence classifier are frozen and the DA component's weights are trained on a minor subset of the labelled target dataset: $\mathcal{L}_{\mbox{\tiny{cross-entropy}}}$ is backpropagated \cite{backprop} to the DA component during the process. Hence, the $D_{KL}(P_S||P_T)$ in Equation~\eqref{eq:cross_entropy} or expressed via the $D_W(P_S, P_T)$ in Equation~\eqref{eq:approx_wasserstein_distance} gets minimized.
\end{enumerate}

\section{RESULTS} \label{sec:results}

\noindent We perform experiments to validate our divergence metric and DA for sEMG-based gesture recognition in case of inter-session and inter-subject scenarios. We follow the exact same hyperparametrization and network implementations as in \cite{2SRNN}. The parameters $\mathbf{M}, \mathbf{M_1}, \mathbf{M_2} \in \mathcal{R}^{f \times f}$ in Equations~\eqref{eq:L-2SRNN} and \eqref{eq:D-2SRNN}, where $f$ is equal to the size of the input features (number of sEMG channels). The $\sigma()$ non-linearity in Equation~\eqref{eq:D-2SRNN} is the REctified Linear Unit \cite{relu}.

For the sequence classifier we use a 2-stack RNN with LSTM cells. Each LSTM cell has a dropout with the probability of 0.5 and 512 hidden units. The RNN is followed by a $G$-way fully-connected layer with 512 units (dropout with a probability of 0.5) and a softmax classifier. Adam \cite{adam} with the learning rate of 0.001 is used for the stochastic gradient descent optimization. The size of the DA component in both the linear and deep cases is less than $1\%$ of the total trainable network parameters. The gesture recognition accuracy is calculated as given below:
\begin{equation}
  \label{eq:classification_accuracy}
  \mbox{Classification Accuracy} = \dfrac{\mbox{Correct}}{\mbox{Total}}*100\%
\end{equation}

We investigate the inter-session and inter-subject divergences and validate our DA method on public sparse and dense sEMG datasets. We follow the experiment setups of previous works for comparability. Since we do sequential modeling in all experiments, we decompose the sEMG signals into small sequences using the sliding window strategy with overlapped windowing scheme. The sequence length must be shorter than 300~ms to satisfy real-time usage constraints. To compare our current experiments with previous works, we follow the segmentation strategy in previous studies.

The dense-electrode sEMG CapgMyo dataset has been thoroughly analysed by \cite{Du2017,10.1371/journal.pone.0206049,2SRNN} such as the sparse-electrode sEMG NinaPro dataset by \cite{ICPR-2014-PatriciaTC,Du2017,2SRNN}.

The CapgMyo dataset \cite{Du2017}: includes HD-sEMG data for 128 electrode channels. The sampling rate is 1 KHz:
\begin{enumerate}
	\item DB-b: 8 isometric, isotonic hand gestures from 10 subjects in two recording sessions on different days.
	\item DB-c: 12 basic movements of the fingers were obtained from 10 subjects.
\end{enumerate}
We downloaded the pre-processed version from \url{http://zju-capg.org/myo/data} to work with the exact same data as \cite{Du2017,2SRNN} for fair comparison. In that version, the power-line interference was removed from the sEMG signals by using a band-stop filter (45–55 Hz, second-order Butterworth). Only the static part of the movements was kept in it (for each trial, the middle one-second window, 1000 frames of data). They used the middle, one-second data to ensure that no transition movements are included in it. We rescaled the data to have zero mean and unit variance, then we rectified it and applied smoothing (as low-pass filtering).

The NinaPro DB-1 dataset \cite{ICPR-2014-PatriciaTC} contains sparse 10-channel sEMG recordings:
\begin{enumerate}
	\item Gesture numbers 1–12: 12 basic movements of the fingers (flexions and extensions). These are equivalent to gestures in CapgMyo DB-c.
\end{enumerate}
The data is recorded at a sampling rate of 100 Hz, using 10 sparsely located electrodes placed on subjects' upper forearms. The sEMG signals were rectified and smoothed by the acquisition device. We downloaded the version from \url{http://zju-capg.org/myo/data/ninapro-db1.zip} to use the exact same data as \cite{Du2017,2SRNN} for fair comparison. For each trial, we used the middle 1.5-second window, 180 frames of data to get the static part of the gestures.

\subsection{Divergence Metric Validation} \label{ssec:result_metric}
\begin{figure}[tbp]
\centerline{\includegraphics[width=\columnwidth,keepaspectratio]{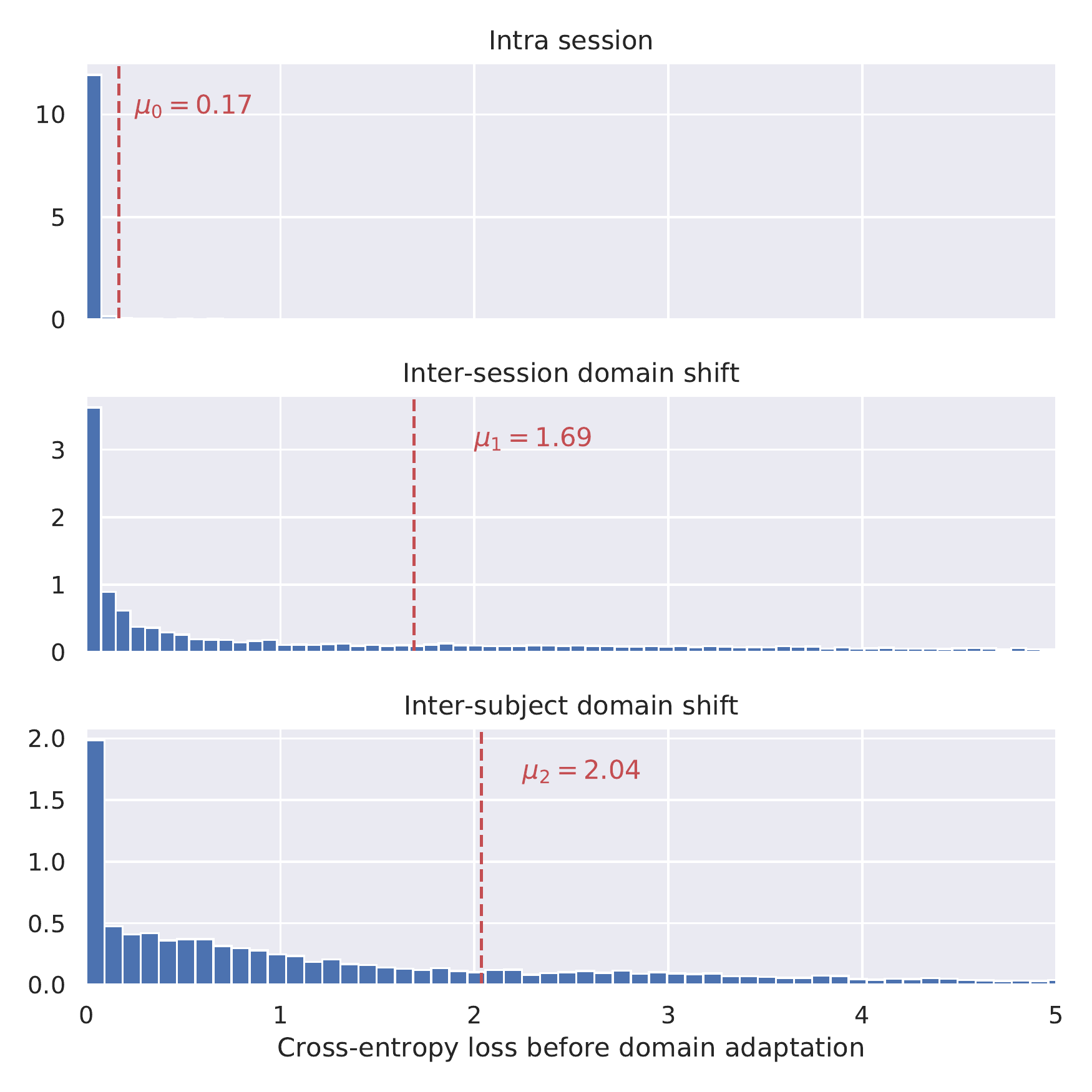}}
\caption{Histograms and mean values in case of the different domain shifts: histograms represent the distributions cross-entropy loss of the classifier on validation data before doman adaptation. The values with red (\textcolor{red}{$\mu_0, \mu_1, \mu_2$}) represent the means of validation loss; \textcolor{red}{$\mu_0$} is the low mean loss in case of intra session; \textcolor{red}{$\mu_1$} and \textcolor{red}{$\mu_2$} (along with their histograms) show high inter-session and inter-subject distribution divergences.}
\label{fig:divergences_pre-DA}
\end{figure}
\begin{figure}[tbp]
\centerline{\includegraphics[width=\columnwidth,keepaspectratio]{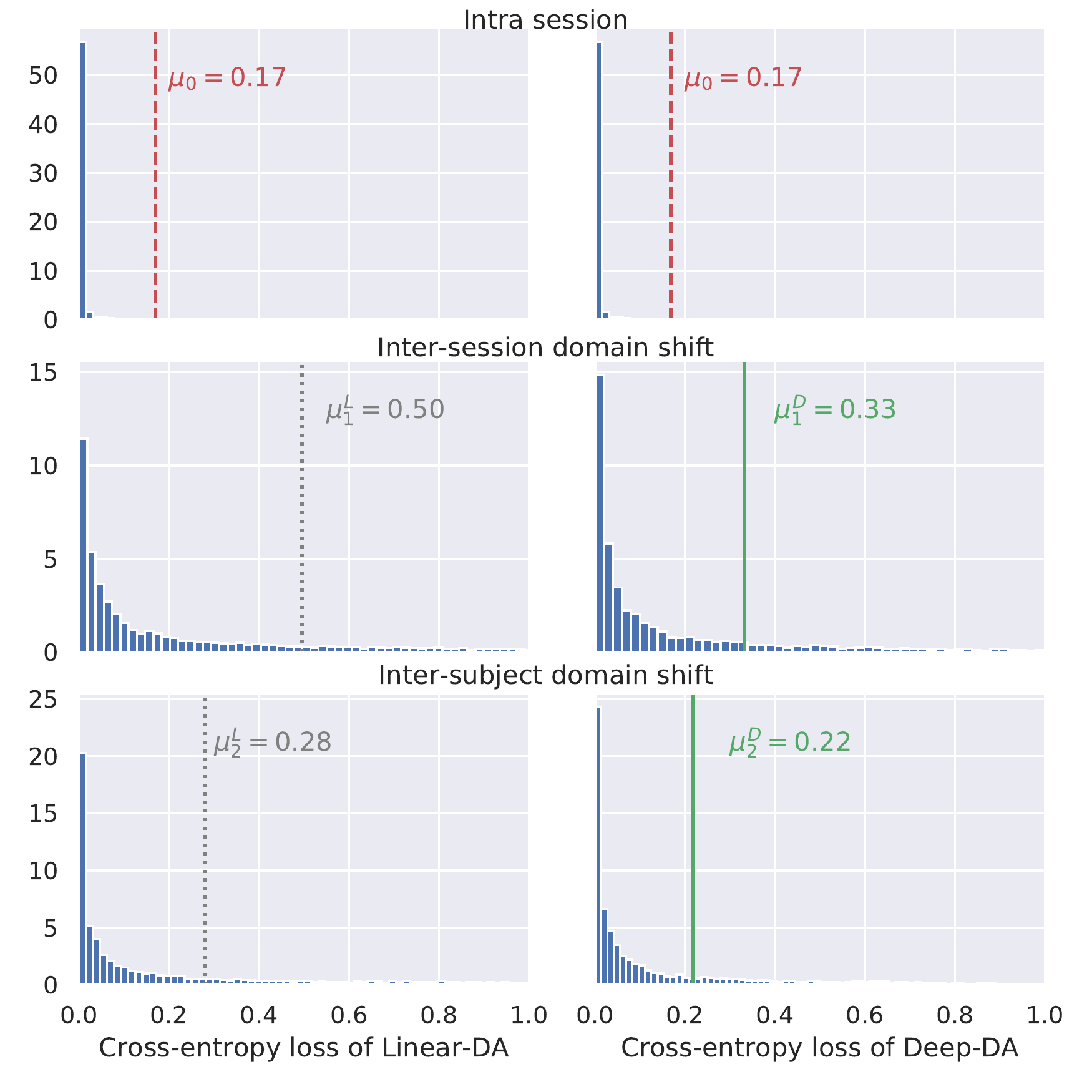}}
\caption{Histograms and mean values in case of different domain shifts and adaptation solutions: histograms represent the distributions of cross-entropy loss of the classifier on validation data after doman adaptation. Intra session statistics (\textcolor{red}{$\mu_0$}) represent the source distribution (towards the divergent distributions are aimed to be adapted). The histograms and corresponding mean values with gray (\textcolor{mygray}{$\mu_1^L, \mu_2^L$}) represent the validation loss after linear domain adaptation; the histograms and corresponding mean values with green (\textcolor{green}{$\mu_2^D, \mu_2^D$}) represent the validation loss after deep domain adaptation.}
\label{fig:divergences_post-DA}
\end{figure}
We validate the proposed domain divergence metric in Section~\ref{ssec:our_divergence_metric} on the CapgMyo DB-b dataset which covers both the inter-session and inter-subject scenarios.

The divergence results are shown in Figure~\ref{fig:divergences_pre-DA} and Figure~\ref{fig:divergences_post-DA}. In both figures the empirical distributions of $\mathcal{L}_{\mbox{\tiny{cross-entropy}}}$ are illustrated by their histograms and mean values.

Figure~\ref{fig:divergences_pre-DA} presents the inter-session and inter-subject divergences before DA. The values with red (\textcolor{red}{$\mu_0, \mu_1, \mu_2$}) represent the means of $\mathcal{L}_{\mbox{\tiny{cross-entropy}}}$; \textcolor{red}{$\mu_0$} is the low mean loss in case of intra session; \textcolor{red}{$\mu_1$} and \textcolor{red}{$\mu_2$} (along with their histograms) show high inter-session and inter-subject domain shifts. \textcolor{red}{$\mu_0$} shows the power of the sequence classifier: \textcolor{red}{$\mu_0$} $=0.17$ is close to the theoretical lower bound of cross-entropy which is $H(S)=0.125$ in the current case.

Figure~\ref{fig:divergences_post-DA} presents the inter-session and inter-subject divergences after DA. Intra session statistics (\textcolor{red}{$\mu_0$}) represent the source distribution (towards the divergent distributions are aimed to be adapted). The histograms and corresponding mean values with gray (\textcolor{mygray}{$\mu_1^L, \mu_2^L$}) represent the validation loss after L-2SRNN DA; the histograms and corresponding mean values with green (\textcolor{green}{$\mu_2^D, \mu_2^D$}) represent the validation loss after D-2SRNN DA. In all cases, the post-DA $\mathcal{L}_{\mbox{\tiny{cross-entropy}}}$ distributions appear to be close to one of the source which is in line with the improved recognition accuracy results in Section~\ref{ssec:result_DA}.

\subsection{Doman Adaptation Validation} \label{ssec:result_DA}

For comparison purposes, we take the exact same pre-trained source classifiers from \cite{2SRNN} and perform D-2SRNN DA (described in Section~\ref{ssec:our_DA}). The evaluation of the D-2SRNN DA is exactly the same as of the L-2SRNN and the AdaBN \cite{Du2017} approaches. Furthermore, the comparison to the MKAL \cite{ICPR-2014-PatriciaTC} also is exactly the same as in \cite{Du2017,2SRNN}.

Table~\ref{table:inter-session} presents the inter-session recognition accuracy results on the dense CapgMyo DB-b dataset. The L-2SRNN and D-2SRNN share the exact same pre-trained source classifier models. The D-2SRNN DA brings $57.3\%$ improvement which is better by $2.1$ percentage points than the L-2SRNN.

Table~\ref{table:inter-subject} shows the inter-subject recognition accuracy results on the dense CapgMyo DB-b \& DB-c and the sparse NinaPro DB-1 datasets. The L-2SRNN and D-2SRNN share the exact same pre-trained source classifier models. The D-2SRNN DA achieves: $74.9\%$ improvement on the DB-b, $156.3\%$ improvement on the DB-c, $107.4\%$ improvement on the DB-1. The performance ratio (between the deep and the linear solutions) is $3.4\%$ in case of the dense datasets, and $11.7\%$ in case of the sparse one which suggests that there is higher gain by non-linear adaptation in case of a sparse-electrode situation.

\begin{table}[tpb]
	\centering
	\begin{threeparttable}
		\setlength{\tabcolsep}{3pt}
		\def\arraystretch{1.5}%
		\caption{Inter-session recognition accuracy results on CapgMyo DB-b.}
		\begin{tabular}{|c|c|c|} 
			\hline
			& pre-DA & post-DA \\
			\hline
			AdaBN \cite{Du2017} & 47.9\% & 63.3\% \\
			L-2SRNN \cite{2SRNN} & 54.6\% & 83.8\% \\
			D-2SRNN & \ditto & \textbf{85.9\%} \\
			\hline
		\end{tabular}
		\label{table:inter-session}
	\end{threeparttable}
\end{table}

\begin{table*}[t]
	\centering
	\begin{threeparttable}
		\setlength{\tabcolsep}{5pt}
		\def\arraystretch{1.5}%
		\caption{Inter-subject recognition accuracy results on dense CapgMyo and sparse NinaPro datasets.}
		\begin{tabular}{|c|c|c|c|c|c|c|} 
			\hline
			& \multicolumn{3}{|c|}{pre-DA} & \multicolumn{3}{|c|}{post-DA} \\
			\hline
			& DB-b & DB-c & DB-1 & DB-b & DB-c & DB-1 \\
			AdaBN \cite{Du2017} & 39.0\% & 26.3\% & - & 55.3\% & 35.1\% & - \\
			MKAL \cite{ICPR-2014-PatriciaTC} & - & - & 30\% & - & - & 55\% \\
			L-2SRNN \cite{2SRNN} & 52.6\% & 34.8\% & 35.1\% & 89.9\% & 85.4\% & 65.2\% \\
			D-2SRNN & \ditto & \ditto & \ditto & \textbf{92.0\%} & \textbf{89.2\%} & \textbf{72.8\%} \\
			\hline
		\end{tabular}
		\label{table:inter-subject}
	\end{threeparttable}
\end{table*}

\section{CONCLUSIONS} \label{sec:conclusions}

\noindent We showed that the divergences between the empirical distributions of the cross-entropy losses by a source classifier trained on the source distribution and evaluated on the target one is a valid measure for the domain shifts between source and target. It works in the absence of source data and a domain adaptation method built on minimizing that divergence is an effective solution in the transductive learning setting. Furthermore, we pointed out that this metric and the corresponding adaptation method is applicable to investigate and improve sEMG-based gesture recognition performance in inter-session and inter-subject scenarios under severe domain shifts. The proposed deep/non-linear transformation component enhances the performance of the 2SRNN architecture especially in a sparse sEMG setting.

The code is available at \url{https://github.com/ketyi/Deep-2SRNN}.

\bibliographystyle{apalike}
{\small
\bibliography{divergence-metrics_and_adapatation-methods}}

\end{document}